\documentclass{article}

\usepackage[final]{nips_2017}

\usepackage[utf8]{inputenc} 
\usepackage[T1]{fontenc}    
\usepackage{hyperref}       
\usepackage{url}            
\usepackage{booktabs}       
\usepackage{amsfonts}       
\usepackage{nicefrac}       
\usepackage{microtype}      

\usepackage{algorithm}
\usepackage{algpseudocode}
\usepackage{url}

\usepackage{amsmath}
\usepackage{amsfonts}       
\usepackage{amssymb}
\usepackage{amsthm}
\usepackage{ifthen}

\usepackage{subfigure} 
\usepackage{graphicx} 

\usepackage{natbib}

\newtheorem{define}{Definition}

\newtheorem{theorem}{Theorem}

\newcommand{\Algorithm}{Situationally Aware oPtions}
\newcommand{\Alg}{SAP}

\newcommand{\xmark}{$\times$}

\title{Situationally Aware Options}

\author{
  Daniel J. Mankowitz \\
  Department of Electrical Engineering\\
  Technion\\
  Haifa, Israel\\
  \texttt{danielm@campus.technion.ac.il} \\
   \And
   Aviv Tamar \\
  Department of Electrical Engineering and
Computer Sciences\\
UC Berkeley\\
Berkeley, CA 94720\\
\texttt{avivt@berkeley.edu}\\
   \AND
   Shie Mannor \\
  Department of Electrical Engineering\\
  Technion\\
  Haifa, Israel\\
  \texttt{shie@ee.technion.ac.il} \\
}

\begin{document}

\maketitle

\begin{abstract}
Hierarchical abstractions, also known as options -- a type of temporally extended action \citep{Sutton1999} that enables a reinforcement learning agent to plan at a higher level, abstracting away from the lower-level details. In this work, we learn \textit{reusable} options whose parameters can vary, encouraging different behaviors, based on the current situation. In principle, these behaviors can include vigor, defence or even risk-averseness. These are some examples of what we refer to in the broader context as Situational Awareness (SA). We incorporate SA, in the form of \textit{vigor}, into hierarchical RL by defining and learning situationally aware options in a Probabilistic Goal Semi-Markov Decision Process (PG-SMDP). This is achieved using our Situationally Aware oPtions (SAP) policy gradient algorithm which comes with a theoretical convergence guarantee. We learn reusable options in different scenarios in a RoboCup soccer domain (i.e., winning/losing). These options learn to execute with different levels of vigor resulting in human-like behaviours such as `time-wasting' in the winning scenario. We show the potential of the agent to exit bad local optima using reusable options in RoboCup. Finally, using SAP, the agent mitigates feature-based model misspecification in a Bottomless Pit of Death domain.
\end{abstract}


\section{Introduction}
\label{sec:intro}

Hierarchical-Reinforcement Learning (H-RL) is an RL paradigm that utilizes hierarchical abstractions to solve tasks. This enables an agent to abstract away from the lower-level details and focus more on solving the task at hand. Hierarchical abstractions have been utilized to naturally model many real-world problems in machine learning and, more specifically, in RL. This includes high-level controllers in robotics \citep{Peters2008,Hagras2004,daSilva2012}, strategies (such as attack and defend) in RoboCup soccer  \citep{Bai2015} and  video games \citep{Mann2015a}, as well as high-level sub-tasks in search and rescue missions \citep{Liu2015}. 
In RL, hierarchical abstractions are typically referred to as skills, \citep{daSilva2012}, Temporally Extended Actions (TEAs), options \citep{Sutton1999} or macro-actions, \citep{Hauskrecht1998a}. We will use the term option to refer to hierarchical abstractions from here on in.
H-RL is important as it utilizes options to both speed up the convergence rate in RL planning algorithms \citep{Mann2013b,Precup1997,Mann2014b} as well as mitigating model misspecification \citep{Mankowitz2016a,Mankowitz2016b,Bacon2015}. While many forms of model misspecification exist in RL \citep{white1994markov,givan1997bounded,nilim2005robust}, we focus on \textit{feature-based} model misspecification which is defined as having a \textit{limited, sub-optimal feature set} (e.g., due to limited memory resources or sub-optimal feature selection) leading to sub-optimal performance.

An important factor missing in H-RL is \textit{Situational Awareness (SA)}. We refer to SA as the ability of an option to be \textit{reused} in different scenarios and vary its behavior according to an \textit{awareness criterion}. Some examples of awareness criteria include (1) Vigor (i.e., force), e.g., an autonomous vehicle is transporting a passenger to a meeting. The agent varies its driving option's level of vigor (for example, increasing/decreasing acceleration) depending on whether the passenger is late/early for his/her meeting; (2) Defence, e.g., an agent is in a dangerous scenario. It has an escape option that can vary its fight-or-flight response depending on the level of danger it is in. (3) Risk, e.g., an agent is buying and selling stocks. Its trading strategy (option) is risk-averse or risk-seeking depending on the current stock market situation and the performance so far in the trading year.

SA can be viewed as being analogous to the level of dopamine in the human brain affecting a human's behavioural responses \citep{Pessiglione2006}. Here, increasing or decreasing the level of dopamine results in complex behaviours that liberate or inhibit the action selection capabilities of the human respectively. This includes controlling the level of vigor, \citep{Salamone2016,Niv2007}, defence \citep{Dayan2012,Lloyd2016,Wenzel2014} and risk-awareness \citep{Clark2014} with which the actions are executed.

In principle, our work has the potential to model a multitude of behaviours. Examples include defence, aggression, risk-awareness and robustness. However, in this paper our focus is on the \textbf{vigor} awareness criterion.  Vigor is defined as the strength, force or rate with which an action (option) is executed \citep{Niv2007}. Therefore, we refer to SA from here on in as (but not limited to) the ability of a reusable option to vary its level of \textit{vigor} based on the current task that the agent is trying to solve. SA can be dependent on time, space, endogenous factors (internal state) and exogenous factors. In this work, we focus on \textbf{time-based} SA.




\begin{figure}[h!]
\centering
\includegraphics[width=1.0\textwidth]{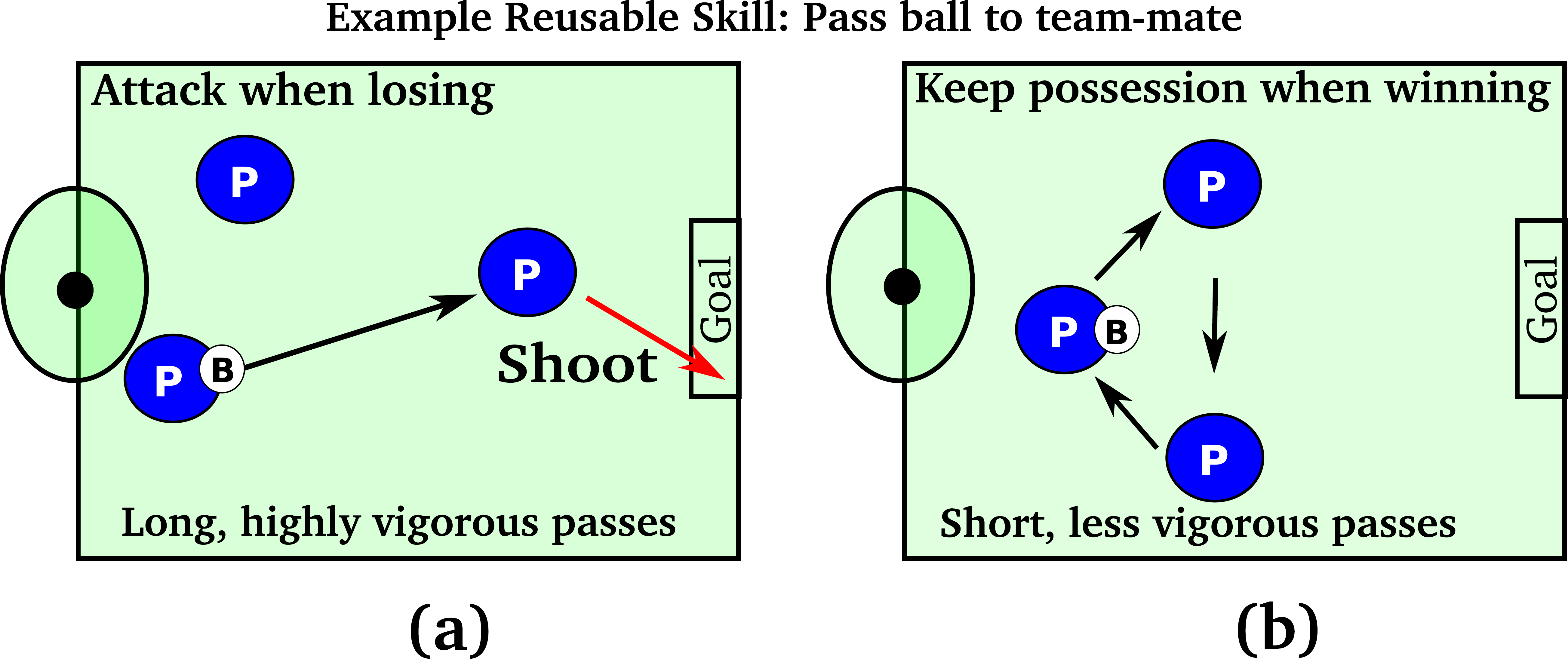}
\caption{Time-based SA - Blue players are on the same team: ($a$) Playing attacking soccer when losing a game and time is running out;  ($b$) Keeping possession and time-wasting when winning the game and time is running out.}
\label{fig:tsa}
\end{figure}

\textbf{Time-based SA:} Consider a soccer game composed of complicated strategies (options), such as attack and defend, which are utilized based on the status of the game. Consider a team losing by one goal to zero with ten minutes remaining. Here, the team needs to play \textit{attacking, highly vigorous} soccer such as dribbling opponents, making long, risky passes as well as shooting from distance to try and score goals and win the game (Figure \ref{fig:tsa}$a$). On the other hand, if the team is winning by one goal to zero with ten minutes remaining, the team needs to `waste time' by maintaining possession and playing \textit{less vigorous, defensive} football (e.g., less dribbling, shorter passes) to prevent the opponent from gaining the ball and scoring goals (Figure \ref{fig:tsa}$b$). In both scenarios the team has the same objective which is to score more goals than their opponent once time runs out (i.e. win the game). In this example, time-based SA enables the agent to \textit{reuse} options such as dribbling and passing across both scenarios, and simply vary each option's level of vigor (e.g., passing strength) based on the amount of time remaining in the task.

%

\begin{figure}
\centering
\includegraphics[width=1.0\textwidth]{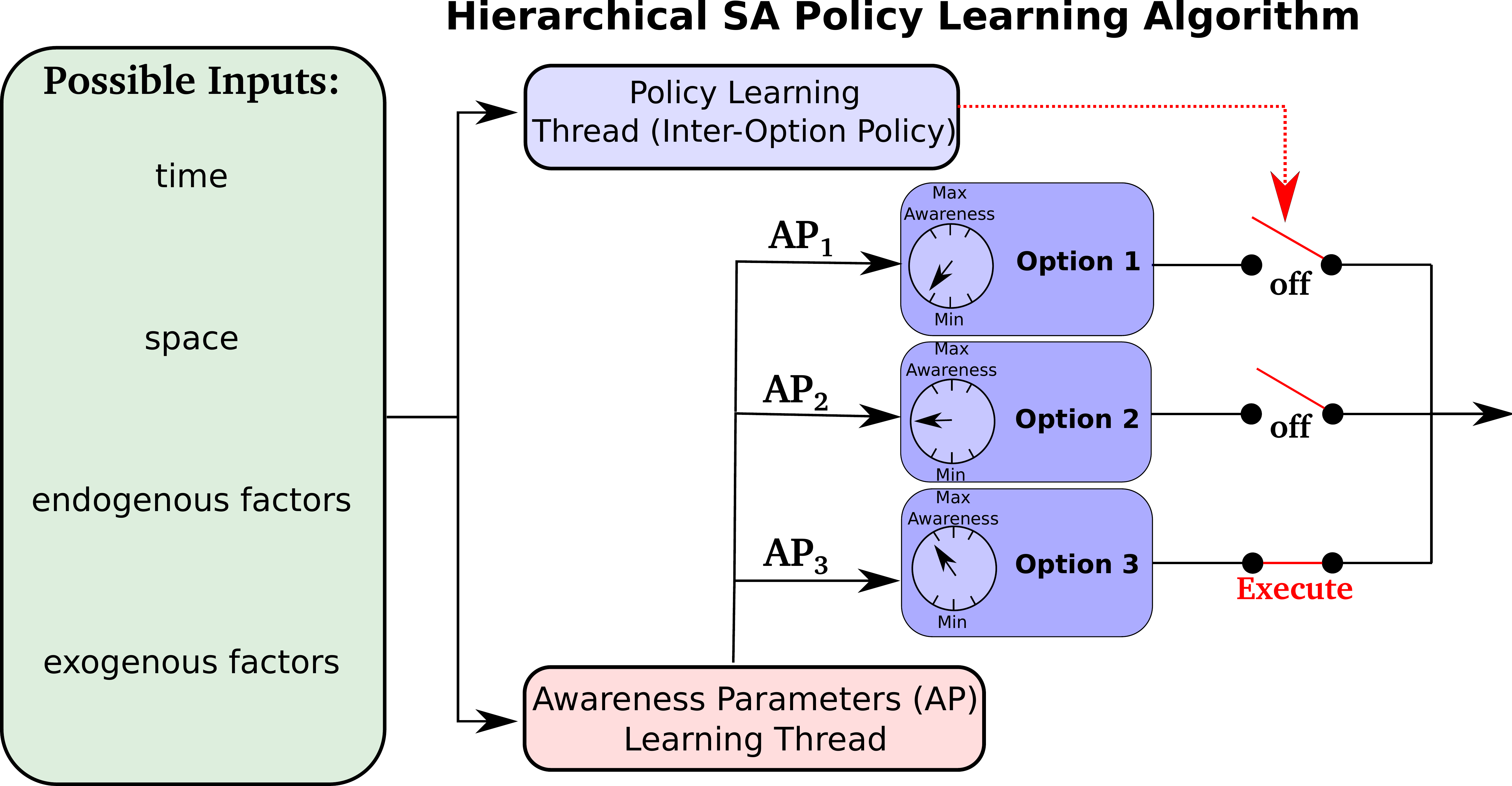}
\caption{Learning a hierarchical SA policy by optimizing the awareness parameters and the policy in separate parallel threads.}
\label{fig:separaterisk}
\end{figure}

Our main idea in this chapter is that a simple way to add SA to H-RL is by maximizing an objective that implicitly captures SA rather than the regular expected return formulation. Previous works have utilized similar objectives, for the purpose of learning risk-aware as well as aggressive policies, in a \textbf{non-hierarchical} setting \citep{Avila1998,Tamar2015a,Tamar2015b,Leibo2017}. Our framework enables an agent, for the first time, to solve a task by incorporating SA into a H-RL objective by learning reusable options with varying levels of SA (see Table \ref{tbl:tea_learning_comparison}). In addition, these reusable SA options can be shown to mitigate feature-based model misspecification.

To induce SA behavior, we incorporate an \textit{awareness parameter} (in our case a vigor parameter) into each option. We 
propose an algorithm that optimizes a SA H-RL objective by learning the awareness parameters (for each option) and the hierarchical RL policy (i.e., inter-option policy) simultaneously in \textit{separate} parallel threads as shown in Figure \ref{fig:separaterisk}. The system receives as input the state, which may include time, space, endogenous or exogenous factors. The resulting learning procedure yields a situationally aware hierarchical RL policy where the awareness parameters control the vigor of the options and therefore implicitly control the vigor of the hierarchical policy. This optimization approach has been previously utilized to encourage risk and safety behaviours in non-hierarchical settings which include autonomous driving \citep{Kuindersma2013,Newman2007}.

In general, many different types of objectives exist that can capture this notion in RL \citep{Borkar2001,Castro2012,Chow2015,Xu2011}. One example SA RL objective that we use in this work is that of a Probabilistic Goal Markov Decision Process (PG-MDP) \citep{Xu2011}. We define a Probabilistic Goal \textit{Semi}-Markov Decision Process (PG-SMDP) which naturally models our \textbf{hierarchical} setting. We optimize the PG-SMDP objective using our Situationally Aware oPtion algorithm (\Alg) by learning the awareness parameters and the hierarchical policy in separate parallel processes. This yields a SA hierarchical policy with learned SA Options (SAOs). 

We reuse the exact same options in different scenarios and \Alg\ learns to vary their levels of vigor to solve the relevant task. We show that the learned SASs exhibit interesting human-like behaviours such as \textit{time-wasting} in a soccer game. However, we do acknowledge that human behaviours are far more complex than those shown in the paper. We also show in our experiments that these options can be used to exit bad local optima as well as overcome feature-based model misspecification, in contrast to the regular expected return formulation.


\begin{table*}\tiny
\centering
\caption{Comparison of approaches to \Alg}
\label{tbl:tea_learning_comparison}
\small
\begin{tabular}{|c|c|c|c|c|c|}
\hline 
 & Maxmizes a  & Learns  & Learns  & Learns  & Mitigate\tabularnewline
 & hierarchical & RL  & \textbf{Reusable} & \textbf{Reusable}  & \textbf{feature-based} \tabularnewline
 & \textbf{SA}  RL& options &  RL & RL Options & model \tabularnewline
& RL objective &  &  with \textbf{SA}& RL Options with \textbf{SA}&  misspecification\tabularnewline
\hline 
\hline 
\Alg\ (this paper) & \checkmark & \checkmark & \checkmark & \checkmark & \checkmark\tabularnewline
\hline 
\citep{Mankowitz2016b} & \xmark & \checkmark & \checkmark & \xmark & \xmark\tabularnewline
\hline 
\citep{Mankowitz2016a} & \xmark & \checkmark & \xmark & \xmark & \xmark\tabularnewline
\hline 
\citep{Bacon2015} & \xmark & \checkmark & \xmark & \xmark & \xmark\tabularnewline
\hline 
\citep{daSilva2012} & \xmark & \checkmark & \checkmark & \xmark & \xmark\tabularnewline
\hline 
\end{tabular}

\end{table*}


\textbf{Main Contributions: } ($1$) Extending H-RL to incorporate SA by defining a PG-SMDP ($2$) The development of the \textbf{S}ituationally \textbf{A}ware o\textbf{P}tions algorithm (\Alg) which optimizes a H-RL SA objective and learns SA Options (SAOs). ($3$) Theorem $1$ which derives a policy gradient update rule for learning SA options and inter-option policy parameters in a PG-SMDP. Theorem $2$ which proves that \Alg\ converges to a locally optimal solution. ($4$) Experiments in the RoboCup domain that exhibit an agent's ability to reuse options while utilizing \Alg\ to vary the options' levels of vigor to solve different soccer scenarios (i.e., winning/losing). (5) Experiments in RoboCup where \Alg\ enables the agent to exit bad local optimal due to sub-optimal reward shaping. In a Bottomless Pit of Death (BPoD) domain \Alg\ shows the ability of the agent to mitigate feature-based model misspecification. Both of these experiments are compared to regular expected return policy gradient baselines.

\section{Background}
\label{sec:background}
\textbf{Semi-Markov Decision Process (SMDP)}: \citep{Sutton1999} A Semi-Markov Decision Process can be defined by the $5$-tuple $\langle S, \mathcal{O}, P, R, \gamma \rangle$, where $S$ is a set of states, $\mathcal{O}$ is a set of options, $P$ is a transition probability function, $\gamma \in [0,1]$ is the discount factor and $R$ is a bounded reward function. We assume that the rewards we receive at each timestep are bounded between $[0, R_{max}]$. Therefore $R$ forms a mapping from $S \times \mathcal{O}$ to $[0, \frac{R_{max}}{1-\gamma}]$ and represents the expected discounted sum of rewards that are received from executing option $o \in \mathcal{O}$ from state $s \in S$. The inter-option policy $\mu:S\rightarrow\Delta_{\mathcal{O}}$ maps states to a probability distribution over options. The goal in an SMDP is to find the optimal inter-option policy $\mu^*$ that maximizes the value function $V^{\mu}(s) = \mathbb{E}\biggl[\sum_{t=0}^{\infty} \gamma^t r_t \vert s, \mu \biggr]$. This represents the expected return of following the inter-option policy $\mu$ from state $s$. The optimal policy $\mu^*$ determines the best action to take for a given state and generates the optimal value function $V^{\mu^{*}}(s)$. 

\textbf{Option, Skill and Macro-Action} \citep{Sutton1999,daSilva2012}: An RL option, skill or macro action $o$ is defined as the $3$-tuple $o= \langle I, \pi_{\theta}, \beta(s) \rangle$ where $I$ is a set of initiation states from which an option can be initialized or executed; $\pi_{\theta}$ is the intra-option policy, which determines the option's behaviour, and is parameterized by $\theta \in \mathbb{R}^n$; The termination probability $\beta(s)$ determines the probability of the option terminating when in state $s$. 

\textbf{Probabilistic Goal MDP (PG-MDP)} \citep{Xu2011}: In some problem domains, optimizing objectives other than the expected return, may be more appropriate. In particular, objectives that maximize the probability of success, and not just the expected outcome, are natural objectives in domains such as finance and operations research, but also in game-playing, such as soccer. The PG-MDP is an extension of the MDP that accounts for such an objective. In a PG-MDP, the goal is to learn a policy $\pi$ that maximizes the probability that some performance threshold will be attained. That is, it aims to maximize $\mathbb{P}(\eta_{\pi} \geq \zeta)$ where $\eta_{\pi}$ is a random variable representing the total reward of the MDP under the policy $\pi$. The parameter $\zeta \in \mathbb{R}$ is a user-defined performance threshold.

\textbf{Policy Gradient} \citep{Peters2006}: In continuous as well as high-dimensional MDPs, it is computationally inefficient to learn a policy that determines an action to perform for any given state. Policies therefore need to be \textit{generalizable}, where the policy will choose the same or similar action to perform when in nearby states. In order to achieve this generalization, a policy is parameterized using techniques such as Linear Function Approximation (LFA) (which we use in this work) \citep{Sutton1998}. A popular technique to learning the parameters for these parameterized policies is the policy gradient method. Let $J(\pi_\theta) = \int_{\tau_{s}} P(\tau_{s}\vert \theta)R(\tau_{s})d\tau_{s}$ denote the expected return of the policy parametrized by $\theta$, where $\tau_{s}$ is a trajectory of $T$ timesteps $\langle s_1,a_1,r_1,s_2 \cdots , s_T \rangle$; $P(\tau_{s}\vert \theta)$ is the probability of a trajectory and $R(\tau_{s})$ is defined as the total reward of the trajectory. Policy gradient uses sampling to estimate the gradient $\nabla_{\theta} J(\pi_\theta)$ and then updates the parameters using a stochastic gradient ascent update rule $\theta_{t+1} = \theta_t + \epsilon \nabla_{\theta} J(\pi_\theta)$ 
where $\epsilon$ denotes a positive step size.

\section{Probabilistic Goal SMDP (PG-SMDP)}
\label{sec:objective}
In this work we focus on solving problems in which the agent must maximize its probability of success for solving a given task in a limited amount of time. A natural model for such problems is the PG-MDP framework described above. However, we are interested in complex problems that require some hierarchical reasoning, and therefore propose to extend PG-MDPs to incorporate options, leading to a PG-SMDP model. We now derive an equivalent PG-SMDP with an augmented state space and option set $\mathcal{O}$ that can easily be utilized with policy gradient algorithms. 

We assume that we are given a set of options $\mathcal{O}= \{ o_i \vert i=1,2, \cdots n, o_j =\langle I_j, \pi_j, \beta_{j}(s) \rangle \}$ and \textit{inter-option policy} $\mu(o \vert s)\rightarrow \Delta_{\mathcal{O}}$ which chooses an option to execute given the current state $s\in S$. We wish to maximize the probability that the total accumulated reward, $\sum_{t=0}^T r_t$, attained during the execution of the \textit{inter-option policy} $\mu$, passes the pre-defined performance objective threshold $\zeta \in \mathbb{R}$ within $T$ timesteps.
This takes the form of the PG-SMDP objective defined in Equation \ref{eqn:probmax}.

\begin{equation}
\max_{\mu} \mathbb{P}(\sum_{t=0}^T r_t \geq \zeta | \mu) \enspace .
\label{eqn:probmax}
\end{equation}

In order to solve this PG-SMDP using traditional RL techniques, we augment the state space with the total \textit{accumulated} reward \citep{Xu2011} to create an equivalent augmented PG-SMDP. We will show the important developments of this formulation for reader clarity. This will enable us to utilize traditional RL techniques in order to maximize the probability of surpassing the performance threshold $\zeta$, given a set of options $\mathcal{O}$, within $T$ timesteps. First note that maximizing the probability can be formulated as an expectation as shown in Equation \ref{eqn:exp}.

\begin{eqnarray}\nonumber
& \max_{\mu} \mathbb{P} \left ( \left. \sum_{t=0}^T r(s_t,o_t) \geq \zeta \right| \mu\right) \\
= & \max_{\mu} \mathbb{E}^\mu \left [ \mathbb{I} \left( \sum_{t=0}^T r(s_t,o_t) \geq \zeta \right) \right] \enspace ,
\label{eqn:exp}
\end{eqnarray}

where $\mathbb{I}(\cdot)$ is the indicator function. This expectation still contains a constraint. We now formulate an equivalent augmented PG-SMDP that incorporates the $\zeta$ constraint into the reward function. Define an augmented state $z = \left \{ s, \eta \right \}$ where $s \in S$ is the original state space and $\eta = \sum_{t=0}^K r(s_t,o_t)$ is the accumulated reward up until current time $K$. We can then define the transition probabilities in terms of the augmented state $z$ as $P(z' \vert z, o) =\left \{ \{ s', \eta+r(s,o)\}  \mbox{w.p } P(s' \vert s, o) \right \}$. The reward function for this augmented state is then defined as:
\begin{equation}
\tilde{r}(z,o)= 
\begin{cases}
    0,& t < T\\
    0,& t=T, \eta < \zeta \\
    1,& t=T, \eta \geq \zeta
\end{cases}
\label{eqn:rz}
\end{equation}

 Together, the transition probabilities and the reward function forms an equivalent PG-SMDP with an augmented state space $z \in Z$. The new, equivalent PG-SMDP objective is shown in Equation \ref{eqn:bellmanaugmented}. This formulation learns an inter-option policy $\mu$ that maximizes the probability that the total accumulated reward will surpass the performance threshold $\zeta$ within $T$ timesteps.
\begin{equation}
\max_{\mu} \mathbb{E} \left [ \sum_{t=0}^T \tilde{r}(z_t,o_t) \right]
\label{eqn:bellmanaugmented}
\end{equation}

In the next Section, we incorporate an Awareness Parameter (AP) into the typical definition of an option to form an SAO. We then derive a policy gradient algorithm to learn both the \textit{inter}-option policy and the APs  by optimizing the PG-SMDP objective.


\section{Situationally Aware Options}
\label{sec:ras}
We modify the typical definition of an option $o_i$ to include an Awareness Parameter (AP) $c_{{w}_{o_{i}}} \in \mathbb{R}$ which is sampled from an Awareness Distribution (AD) $c_{{w}_{o_{i}}} \sim P_{w_{o_{i}}}(\cdot)$ with parameters ${w_{o_{i}}} \in \mathbb{R}^m$. This is the parameter that controls the SA (in our case, the level of vigor) of the Situationally Aware Option (SAO). 

\begin{define}
A Situationally Aware option $o$ is a temporally extended action that consists of the $4$-tuple $o = \langle I, \pi_{\theta}, \beta(z), c_{{w}_{o}}\rangle$, where $I$ are the set of  states from where the SAO can be initialized; $\pi_\theta$ is the parameterized intra option policy; $\beta(z)$ is the probability of terminating in state $z \in Z$;  and $c_{{w}_{o}} \in \mathbb{R}$ is the Awareness Parameter (AP) governed by the Awareness Distribution (AD) $c_{{w}_{o}} \sim P_{w_{o}}(\cdot)$.
\end{define}

In practice, the AP can parameterize the intra-option policy, or act as a meta-parameter for the SAO (e.g., Dribble power in the RoboCup experiment (See Experiments Section)). 

\section{\Alg\ Algorithm }
\label{sec:sars}
The \Algorithm\ algorithm (\Alg) learns the parameters of a \textit{two-tiered} option selection policy:

\begin{equation}
\mu_{\alpha,\Omega_{i}}(o_i,c| z) = \mu_{\alpha}(o_i|z)\mu_{\Omega_{i}}^{o_{i}}(c|z) \enspace ,
\end{equation}

where $\mu_{\alpha}:Z \rightarrow \Delta_{\mathcal{O}}$ is the \textit{inter-option} policy, parameterized by $\alpha \in \mathbb{R}^d$, that selects which SAO $o_i$ needs to be executed from a set $\mathcal{O}$ of $N$ SAOs, given the current state $z \in Z$;
$\mu_{\Omega_{i}}^{o_{i}}(\cdot \vert z)$ is the AD for SAS $o_i$ with AD parameters $\Omega_i = w_{o_{i}} \in \mathbb{R}^{m}$. The AD parameters for all SASs are stored in a vector $\Omega=[\omega_{o_{1}}, \omega_{o_{2}}, \cdots, \omega_{o_{N}}] \in \mathbb{R}^{|N||m|}$.
 
The two-tiered option selection policy is executed by first sampling a SAO $o_i$ to execute from $\mu_{\alpha}(\cdot|z)$. The situational awareness of the option is then determined by sampling the AP $c_{w_{o_{i}}}$ from the AD $\mu_{\Omega_{i}}^{o_{i}}(\cdot|z)$. \Alg\ learns \textbf{(1)} the inter-option policy parameters $\alpha \in \mathbb{R}^d$ and \textbf{(2)} the AD parameters $\Omega$ to produce SAOs. In order to derive gradient update rules for these parameters in a policy gradient setting, we define the notion of an awareness trajectory.


\textbf{Awareness Trajectory:} In the standard policy gradient framework, we define a typical trajectory as $\tau_s = (z_t, o_t, r_t, z_{t+1})_{t=0}^{T}$ where $T$ is the length of the trajectory. To incorporate the two-tiered policy into this trajectory, we define an awareness trajectory $\tau = (z_t, o_t, c_{w_{o_{t}}} r_t, z_{t+1})_{t=0}^{T}$ where at each timestep, we draw an AP $c_{w_{o_{t}}}$ corresponding to the SAO $o_t$ that was selected. We can therefore define the probability of a trajectory as $\mathbb{P}_{\alpha, \Omega}(\tau) = \mathbb{P}(z_0)\prod_{t=0}^{T-1} \mathbb{P}(z_{t+1} \vert z_t, o_t, c_t) \mu_{\alpha,\Omega}(o_t,c_t| z_t)$, where $P(z_0)$ is the initial state distribution; $P(z_{t+1} \vert z_t, o_t, c_t)$ is the transition probability of moving from state $z_t$ to state $z_{t+1}$ given that a SAO $o_t$ was executed with AP $c_t$; and $\mu_{\alpha,\Omega}(o_t,c_t| z_t)$ is the two-tiered selection policy. Using this notion, we now derive the gradient update rules for $\alpha$ and $\Omega$ as shown in Theorem \ref{thm:grad}.

\subsection{Inter-option policy and AD Update Rules} 
We define the expected reward for following a policy $\mu_{\alpha, \Omega}$ as $J(\mu_{\alpha, \Omega}) = \int_{\tau} P(\tau \vert \alpha, \Omega)R(\tau)d\tau$. Taking the derivative of this objective with respect to $\alpha$ and $\Omega$, using the well-known likelihood trick \citep{Peters2008}, leads to the gradient update rules in Theorem \ref{thm:grad}. The full derivation can be found in the supplementary material.

\begin{theorem}
Suppose that we are maximizing the Policy Gradient (PG) objective $J(\mu_{\alpha, \Omega}) = \int_{\tau} P(\tau\vert \alpha, \Omega)R(\tau)d\tau$ using awareness trajectories, generated by the two-tiered option selection policy $\mu_{\alpha}(o_t|x_t)\mu_{\Omega}^{o_{t}}(c_t|z_t)$, then the gradient update rules for the inter-option policy parameters $\alpha \in \mathbb{R}^d$ and the AD parameters $\Omega \in \mathbb{R}^{|N||m|}$ are defined in Equations \ref{eqn:grad1} and \ref{eqn:grad2} respectively.

\begin{equation}
\nabla_\alpha J(\mu_{\alpha, \Omega}) = \left <  \sum_{h=0}^H  \nabla_\alpha \log  \mu_{\alpha}(o_h|z_h) \sum_{j=0}^H \gamma^j r_j     \right >
\label{eqn:grad1}
\end{equation}
\begin{equation}
\nabla_\Omega J(\mu_{\alpha, \Omega}) = \left <  \sum_{h=0}^H  \nabla_\Omega \log  \mu_{\Omega}^{o_{h}}(c_h|z_h) \sum_{j=0}^H \gamma^j r_j     \right > 
\label{eqn:grad2}
\end{equation}
$H$ is the trajectory length and $<\cdot >$ is an average over trajectories as in standard PG.
\label{thm:grad}
\end{theorem}

Given the gradient update rules, we can derive an algorithm for learning both the inter-option parameters $\alpha$ and the continuous AD parameters  $\Omega$ for the $N$ SAOs. \Alg\ learns these parameters by two timescale stochastic approximation, as shown in Algorithm \ref{alg:sars-rl}, and converges to a locally optimal solution as is proven in Theorem \ref{thm:convergence}. The convergence proof is based on standard two-timescale stochastic approximation convergence arguments  \citep{Borkar1997} and is found in the supplementary material.

\begin{theorem}{(\Alg\ Convergence.)}
Suppose we are optimizing the expected return $J(\mu_{\alpha,\Omega})=\intop P(\tau \vert \alpha,\Omega) R(\tau)d\tau$
for any arbitrary \Alg\ policy $\mu_{\Omega,\alpha}$ where $\alpha\in\mathbb{R}^{d}$ and $\Omega\in\mathbb{R}^{|N||m|}$
 are the inter-option policy and AD parameters respectively. Then, for step sizes sequences
$\{a_{k}\}_{k=0}^{\infty},\{b_{k}\}_{k=0}^{\infty}$ that satisfy the conditions
$\sum_{k}a_{k}=\infty,\sum_{k}b_{k}=\infty,\sum_{k}a_{k}^{2}<\infty,\sum_{k}b_{k}^{2}<\infty$
and $b_{k}>a_{k}$, \Alg\ iterates converge a.s $\alpha_k \rightarrow \alpha^{*},\Omega_k \rightarrow\bar{\lambda}(\alpha^{*})$ as $k \rightarrow \infty$
to the countable set of locally optimal points of $J(\mu_{\Omega,\alpha})$.
\label{thm:convergence}
\end{theorem} 


\begin{algorithm}
\caption{\Alg\ Algorithm}
\label{alg:sb}
\begin{algorithmic}[1]
\Require $\mu_{\alpha, \Omega}$ \Comment{Arbitrary inter-option policy}
\State \textbf{repeat}:
\State Sample trajectory $\tau$ using $\mu_{\alpha_{k}, \Omega_{k}}$
\State Calculate $\nabla_{\alpha} J(\mu_{\alpha_{k}, \Omega_{k}})$ and $\nabla_{\Omega} J(\mu_{\alpha_{k}, \Omega_{k}})$
\State $\alpha_{k+1} \rightarrow \alpha_k + a_k \nabla_{\alpha} J(\mu_{\alpha_{k}, \Omega_{k}})$ 
\State $\Omega_{k+1} \rightarrow \Omega_k + b_k \nabla_{\Omega} J(\mu_{\alpha_{k}, \Omega_{k}})$
\State $\alpha_k = \alpha_{k+1}$, $\Omega_k = \Omega_{k+1}$  \Comment{stepsize $b_k > a_k$}
\State \textbf{until convergence}
\end{algorithmic}
\label{alg:sars-rl}
\end{algorithm}



\section{Experiments}
\label{sec:experiments}
The experiments were performed in (1) a RoboCup 2D soccer simulation domain \citep{Akiyama2014}; a well-known benchmark for many AI challenges and (2) a novel Bottomless Pit of Death domain (BPoD). In the experiments, we demonstrate the ability of the agent to learn to reuse options with varying levels of vigor (SAOs), inducing different behaviours such as `time-wasting', by optimizing the PG-SMDP objective using \Alg. In addition, we show in RoboCup (i)  the potential for \Alg\ to exit bad local optima as a result of sub-optimal reward shaping. In the BPoD domain, we show (ii) the agent's ability to utilize SA to mitigate feature-based model misspecification. In (i) and (ii) \Alg\ is compared to policy gradient algorithms that maximize a regular Expected Return (ER) SMDP objective.

\subsection{RoboCup Offense (RO) Domain} 
\begin{figure*}
\centering
\includegraphics[width=1.0\textwidth]{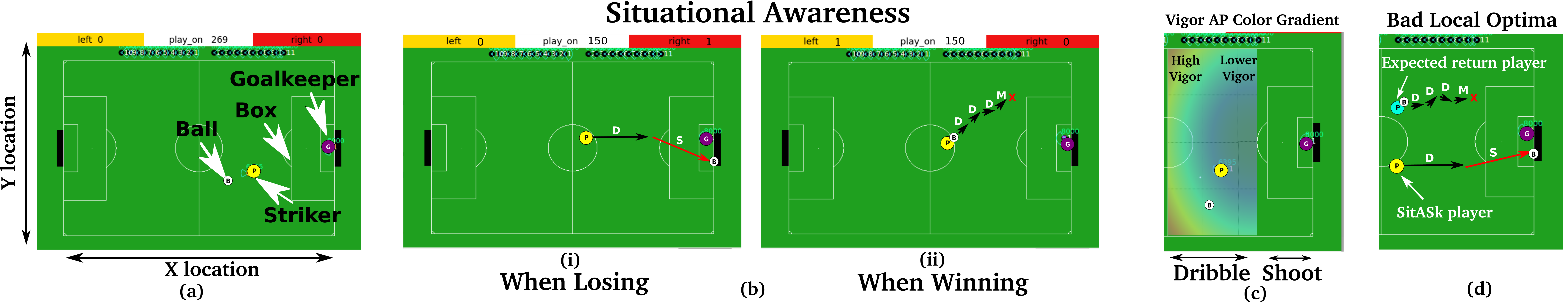}
\caption{($a$) The RO domain ($b$) SA for a losing scenario ($i$) and a winning scenario ($ii$) ($c$) The superimposed vigor AP color gradient for the Dribble option. ($d$) Exiting bad local optima: the trajectories for the trained Expected Return (ER) and \Alg\ policies.}
\label{fig:r1}
\end{figure*}

This domain\footnote{\tiny \url{https://github.com/mhauskn/HFO}} consists of two teams on a soccer field where the striker (the yellow agent) needs to score against a goalkeeper (purple circle) as shown in Figure \ref{fig:r1}$a$. The striker has $T=150$ timesteps (length of the episode) to try and score a goal. \textbf{State space} - $\langle x,y, \eta \rangle$ where $\langle x,y \rangle$ is the continuous field location of the striker and $\eta=\sum_{t=0}^K r_t$ is the sum of accumulated rewards up to current time $K$. \textbf{Options} - The SAO set $\mathcal{O}$ in each of the experiments consists of three SAOs: (1) Move to the ball \textbf{(M)}, (2) Move to the ball and shoot towards the goal \textbf{(S)} and (3) Move to the ball and dribble in the direction of the goal \textbf{(D)}. Each SAO $o_i$ is parameterized with an AP $c_{w_{o_{i}}}$.  We focus on learning the dribbling power AP $c_{w_{D}}$ for the Dribble option that controls how hard the agent kicks the ball (i.e., the level of vigor with which the agent dribbles). \textbf{Data:} \Alg\ is trained over $3$ independent trials with $20,000$ episodes per trial. \textbf{Learning Algorithm and features} - The learning algorithm is an Actor Critic Policy Gradient (AC-PG) version of \Alg\ \footnote{\tiny AC-PG has lower variance compared to regular PG and the convergence guarantees are trivial extensions of the current proof.}. The inter-option policy $\mu$ is represented by a Gibb's distribution with Fourier Features \citep{Konidaris2008}. The AD is represented as a normal distribution $c_{w_{o_{i}}} \sim \mathcal{N}(\phi(s)^T w_{o_{i}}, V)$ with a fixed variance $V$. Here, $\phi(s)=[1,x_{agent},y_{agent},\eta,distGoal]$ where $\langle x_{agent},y_{agent} \rangle$ is the agent's location and \textit{distGoal} is the distance of the agent to the goal. \textbf{Rewards} - Engineering of the reward in RL is common practice for the RoboCup domain \citep{Hausknecht2015,Bai2015}. The rewards for both of the RoboCup scenarios have been engineered based on logical soccer strategies. The striker gets small positive rewards for dribbling outside the box $r_{D,far}$ and shooting when inside or near the box $r_{S,near}$. Negative rewards come about when the striker dribbles inside the box, $r_{D,near}$, or shoots from far, $r_{S,far}$, as the striker has a smaller probability of scoring \citep{Yiannakos2006}. The striker also gets a small positive reward for moving towards the ball $r_{M}$. There is also a game score reward, $r_{score}$, received at each timestep, that is positive if winning and negative if losing or drawing. \textbf{PG-SMDP Setup (for \Alg):} The rewards $\tilde{r}=1$ if $\eta>=\zeta$ at the end of each episode, otherwise the reward is $0$ at each timestep (see Equation \ref{eqn:rz}). \textbf{SMDP Setup (for ER):} In the expected return setting, there is no performance threshold $\zeta$ and the regular rewards are utilized at each timestep.


%



\subsubsection{Learning Situationally Aware optionS}
\label{sec:sa}
In the experiments to follow, we show that learning the inter-option policy and AD parameters using \Alg\ to optimize a PG-SMDP objective can bring about SAOs that exhibit time-based SA. We provide the agent with two different soccer situations: \textbf{(1)} The agent is losing the game $0-1$, and \textbf{(2)}  winning the game $1-0$. Similar results are obtained for different scores (e.g., $2-0$ and $0-2$ etc. and have therefore been omitted). In each scenario, the \textit{same} options are reused, but \Alg\ learns to modify the AD parameters (and therefore the APs) to induce different, situation-specific behaviours with varying levels of vigor. For all of the scenarios, the performance threshold $\zeta$ is set to a constant value ($\zeta=1.0$) \textit{a-priori}. Throughout the RoboCup experiments, we will refer to Table \ref{tab:capt}. In the table, \textit{Goals} refers to the average number of goals (mean$\pm$std) scored per  $100$ episodes. \textit{Out of Time} refers to the average number of times the agent failed to score a goal and ran out of time. \textit{Avg Reward} is the average reward attained by the agent and \textit{Episode Length} refers to the average length (number of timesteps) of each episode.



\textbf{\underline{SA in a Losing Scenario:}} In a scenario where a team is losing and time is running out, the team needs to play aggressive, attacking soccer to try and score goals. The agent is placed in a losing scenario where the score  is $0-1$ to the opposition with $150$ timesteps remaining. Using \Alg\, the agent learns the AP $c_{w_{D}}$ for the Dribble option. The agent learns to perform an aggressive, highly vigorous Dribble by kicking the ball with significant power to make quick progress along the pitch and get in a position to shoot for goal as seen in Figure \ref{fig:r1}$b(i)$. The average AP value for the Dribble SAO is $100$ (max value $150$, min value $0$). The AP is state dependent, causing the agent to initially dribble with great vigor when near the half-way line and then decrease the dribble power when approaching the goal to prevent losing possession to the goalkeeper. This is seen from the learned dribble AP's vigor color gradient superimposed onto the RO domain in Figure \ref{fig:r1}$c$. The color gradient varies from powerful, highly vigorous kicks (in red)  to softer, less vigorous kicks (in blue). Once the agent is near the goal, it executes the option \textbf{Shoot} (Figure \ref{fig:r1}$b(i)$). The average episode length is $70.0\pm1.0$ (mean$\pm$std) as seen in Table \ref{tab:capt} and the average number of goals scored over $100$ evaluation episodes is $74.3\pm6.5$. In addition, the keeper captures the ball on average  $21\pm 5.29$ times indicating that the striker is playing aggressive, highly vigorous football and is consistently shooting for goal. In addition, the average reward is consistently higher than the $\zeta$ threshold.


\textbf{\underline{SA in a Winning Scenario:}} When winning a game with little time remaining, a natural strategy is to hold onto the ball and \textit{run-out-the-clock} (\textbf{`time-wasting'}) so as to prevent the opposing team from gaining possession and possibly scoring a goal. The agent is placed in a scenario where it is winning the game $1-0$. \Alg\ reuses the \textit{same} Dribble option as in the losing scenario but learns a conservative, less vigorous AP which causes the agent to run-out-the-clock. In this case, the agent learns to slowly dribble its way up the pitch, collecting the dribble from far rewards $r_{D,far}$ in the process as seen in Figure \ref{fig:r1}$b(ii)$. Once the agent crosses the performance threshold, it stands on the ball, and \textit{wastes time} by executing the \textbf{M} option, whilst continuing to collect the positive score rewards $r_{score}$ and the positive $r_{M}$ rewards. This strategy causes the agent to take the largest amount of time on average ($142.3\pm1.5$ steps in Table \ref{tab:capt}) to complete each episode (time wasting) and as a result only scores $1.3\pm0.6$ goals. However, the ball is almost never captured by the opponent ($7.3\pm0.6$ times on average), since the agent is playing less vigorous, more conservative football.  A \textbf{Video}\footnote{\tiny\url{https://youtu.be/2T_jYPXsI7k}} of the agent's behavior in each of these scenarios is available online.

\textbf{\underline{Exiting bad local optima using \Alg:}} In the losing scenario, we noticed that the SAOs have the potential to overcome bad local optima due to sub-optimal reward shaping. We compared \Alg, which optimizes a PG-SMDP objective, to an AC-PG Expected Return (ER) algorithm which maximizes a regular SMDP objective (i.e., utilizes the regular rewards to learn a goal-scoring policy). As seen in Figure \ref{fig:r1}$d$, the ER striker (light blue circle) does not learn to score goals, instead choosing to collect the positive rewards $r_{D,far}$ and $r_{M}$.  This causes the ER agent to execute \textbf{D} until it settles on the \textbf{M} option and stands on the ball. As seen in Table \ref{tab:capt}, the ER agent only manages to score $1.7\pm1.2$ goals on average and has a low average reward $-0.3\pm0.1$, well below the $\zeta$ threshold. This behaviour is plausible since policy gradient algorithms tend to converge to local optima very quickly in practice.  Collecting $r_{D,far}$ and $r_{M}$ is not enough for the \Alg\ agent (yellow circle in Figure \ref{fig:r1}$d$) to pass its performance threshold $\zeta$, since there are also negative game score rewards $r_{score}$ attained at each timestep in the losing scenario. This forces the \Alg\ agent to search for additional rewards such as a goal-scoring reward. As seen in Table \ref{tab:capt}, the \Alg\ agent learns to score goals ($74\pm6.5$), and achieves average reward well above the $\zeta$ performance threshold. As a result it avoids falling into the bad local optima unlike the ER simulation.

\begin{table}[h!]
\centering
\caption{Performance of the trained \Alg\ and ER policies in the winning/losing scenarios averaged over $100$ evaluation episodes.}
\small
\begin{tabular}{|c|c|c|c|}
\hline 
 & \textbf{\Alg} & \textbf{\Alg} & \textbf{ER}\tabularnewline
 & Winning & Losing & Losing\tabularnewline
\hline 
\hline 
Goals & $1.3\pm0.6$ & 74.3$\pm$6.5 & 1.7$\pm$1.2\tabularnewline
\hline 
Out of Time & 90.3$\pm$1.5 & $1.0\pm1.7$ & $47.0\pm3.6$\tabularnewline
\hline 
Avg Reward & $3.9\pm1.1$ & 6.3$\pm$0.2 & -0.3$\pm$0.1\tabularnewline
\hline 
Episode Length & $142.3\pm1.5$ & $70.0\pm1.0$ & $107.3\pm3.8$\tabularnewline
\hline 
\end{tabular}
\label{tab:capt}
\end{table}

\begin{figure*}
\centering
\includegraphics[width=1.0\textwidth]{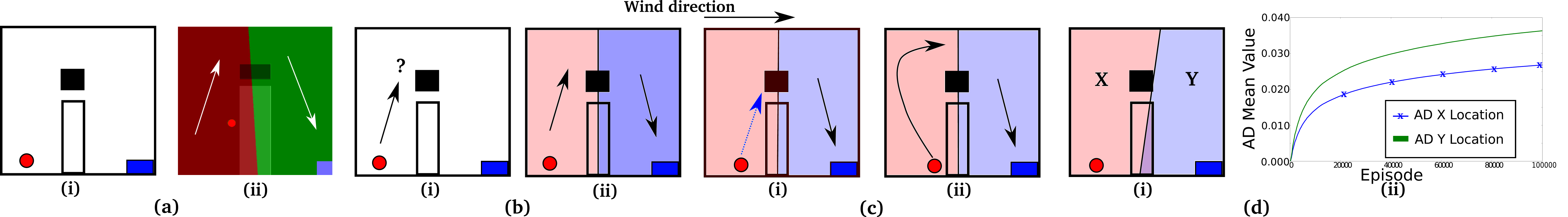}
\caption{($a$): ($i$) The BPoD Domain. ($ii$) The options (white arrows) and options partitions (red and green regions indicating where the options operate) learned using the ASAP algorithm. ($b$)($i$) The policy representation is not rich enough to solve the task. ($ii$) Options mitigate the misspecification by duplicating the policy. ($c$)($i$) Given the wind factor, an agent starting near the wall, will ultimately get forced into the bottomless pit. ($ii$) Mitigating the misspecification may be possible by learning SASs. ($d$): ($i$) The $X$ and $Y$ locations where the AD parameters are evaluated. ($ii$) The learned AD mean values for the $X$ and $Y$ locations.}
\label{fig:b1}
\end{figure*} 

\subsection{Bottomless Pit of Death Domain}
The Bottomless Pit of Death (BPoD) domain contains two rooms, separated by a dividing wall as shown in Figure \ref{fig:b1}$a(i)$. Above the wall is a bottomless pit of death (black square). There is a wind factor in this domain that pushes the agent in an easterly direction, as shown by the arrow, with random velocity. The agent (red ball) needs to circumnavigate both the wall and the pit to reach the goal location (blue square). \textbf{State space} - $\langle x,y, \eta \rangle$ where $\langle x,y \rangle$ is the location of the agent. \textbf{Options} - Options are represented as a probability distribution over four primitive actions (i.e., North, South, East, West), independent of the state. \textbf{Data} - $10$ independent trials of $100,000$ episodes were run for each experiment. \textbf{Rewards} - The agent receives negative rewards ($-1$) for colliding with the wall and after each timestep. A large negative reward ($-10$) is attained if the agent falls into the bottomless pit. A positive reward ($+100$) is received if the agent reaches the goal location. \textbf{Learning Algorithm and features} - The learning algorithm for the APs is an AC-PG version of \Alg. The inter-option policy $\mu$ and the options are initialized using the Adaptive Skills, Adaptive Partitions (ASAP) \citep{Mankowitz2016b} SMDP algorithm. The Awareness Distribution (AD) is represented as a normal distribution $c_{w_{o_{i}}} \sim \mathcal{N}(\phi(s)^T w_{o_{i}}, V)$ with a fixed variance $V$. Here,  $\phi(s)$ are state dependent features $[1,x,y,x^2, y^2, \eta]$ where $\langle x,y \rangle$ represent the agent's location. \textbf{PG-SMDP Setup (for \Alg):} The reward threshold $\zeta=50$ is set \textit{a-priori}. The agent receives a reward of $\tilde{r}=0$ at each timestep and a $\tilde{r}=1$ if $\eta>=\zeta$ by the end of each episode (see Equation \ref{eqn:rz}). \textbf{SMDP Setup (for ER):} No $\zeta$ threshold and the regular rewards are utilized at each timestep.


\begin{figure}[h!]
\centering
\includegraphics[width=1.0\textwidth]{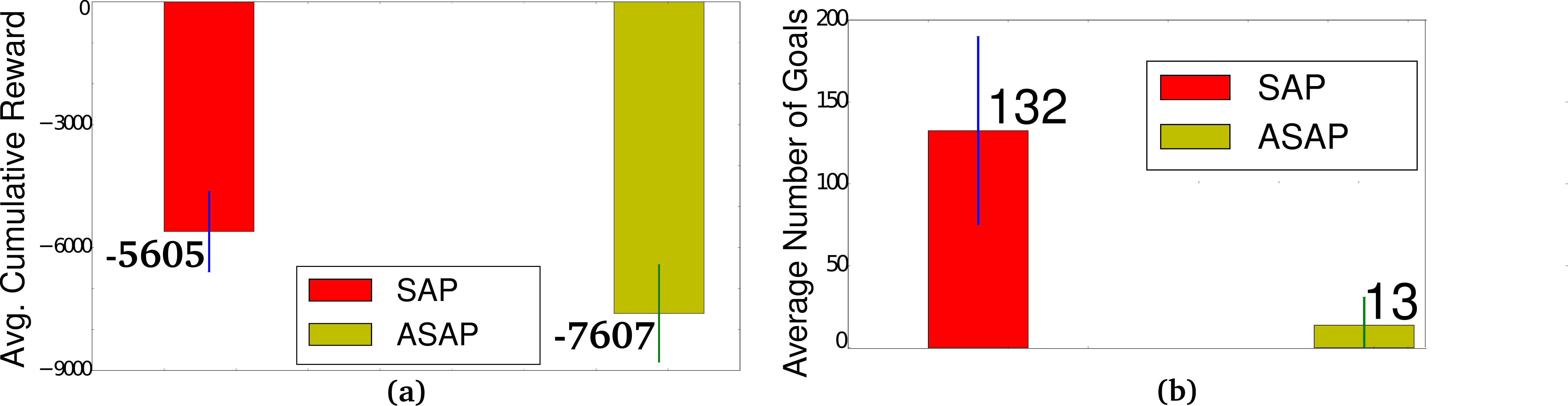}
\caption{($a$) The average cumulative reward performance of \Alg\ compared to the regular ASAP algorithm. ($b$) The average number of goals reached.}
\label{fig:performance}
\end{figure}

\subsubsection{Feature-based Model Misspecification}
\label{sec:featurebased}
In the experiments to follow, we will provide an example in the BPoD domain where SAOs learned using \Alg\ can be used to mitigate feature-based model misspecification. Consider Figure \ref{fig:b1}$b(i)$. Given the limited policy representation of a probability distribution over actions which are independent of the state, the agent is unable to circumnavigate the obstacle and reach the goal region resulting in feature-based model misspecification. Numerous works \citep{Mankowitz2016a,Mankowitz2016b,Bacon2015} have shown that options are necessary to mitigate this form of misspecification. This is achieved  by duplicating the policy representation and generating multiple options that, collectively, can solve the task as seen in Figure \ref{fig:b1}$b(ii)$. The red region is where the optin \textit{move up and to the right} operates and the blue region is where the option \textit{move down and to the goal} operates.




However, even though options may produce a reasonable policy, it is still possible to end up with a misspecified model. If the agent starts in a location near the bottom of the domain on the left side of the wall (Figure \ref{fig:b1}$c(i)$), then the agent will ultimately get forced into the pit due to the wind factor. This misspecification can be mitigated by learning SAOs. 

\underline{\textbf{Learning Options:}} First, we initialize the options, given the limited feature representation, using ASAP. This algorithm learned a set of options (the superimposed white arrows in Figure \ref{fig:b1}$a(ii)$ for visualization purposes) and option partitions (the red and green regions in Figure \ref{fig:b1}$a(ii)$, which define where the options operate). ASAP attempts to mitigate the feature-based model misspecification. However, if the agent is evaluated from the location in Figure \ref{fig:b1}$c(i)$ (the red ball), the agent is forced into the pit by the wind factor and a misspecification is still present. 
To solve this problem, we reuse the same options and incorporate a \textit{vigor} AP into each option to form SAOs. We will show that learning the AP for each option enables the agent to mitigate this misspecification.

\underline{\textbf{Learning Situationally-Aware Options: }} A single AP is added to each option's move East primitive action. The AP takes problem-specific values in the range $[-0.05, 0.05]$. The AP can either increase the magnitude (and therefore the vigor) of the agent's movement in the Easterly direction by applying a positive value. Alternatively, it can decrease, or even change the direction of the agent's movement for negative values. The hope is that \Alg\ learns AP's in the left partition that generate a less vigorous East action (Figure \ref{fig:b1}$c(ii)$), causing the agent to avoid the pit. In the right partition, the AP should have larger positive values to drive the agent with great vigor towards the goal. 

After learning the AP using \Alg, the agent's average performance was evaluated for $1000$ episodes over $10$ independent trials. Figure \ref{fig:b1}$d(ii)$, plots the learned means of the AD parameters, from which the APs are sampled, at two specific locations $X$ and $Y$ (Figure \ref{fig:b1}$d(i)$). As can be seen in the graph, the mean AD value at location $X$ is significantly lower, causing it to generate, on average, more negative AP values which inhibit the move East action and helps the agent avoid the pit. On the other hand, the mean AD value at location $Y$ is larger, causing the agent to vigorously approach the goal location. This causes the agent to mitigate the feature-based misspecification.  \Alg\ is compared to the average performance of the ASAP algorithm (which optimizes a regular ER SMDP objective). Figure \ref{fig:performance}$a$ indicates that \Alg\ generates higher average cumulative reward compared to the regular ASAP agent. In addition, the \Alg\ agent reaches the goal on average $132$ times compared to only $13$ using ASAP (Figure \ref{fig:performance}$b$). 




\section{Discussion}
We have defined a PG-SMDP which provides a natural objective for incorporating SA into hierarchical RL. We have introduced Situationally Aware Options (SAOs) --- a type of parameterized temporally extended action \citep{Sutton1999} with an additional Awareness Parameter (AP). We show that the same options can be \textit{reused} in different scenarios to induce varying behaviours (e.g., time-wasting) based on the AP's level of vigor. We have developed the \Algorithm\ algorithm (\Alg) which learns both the inter-option policy that chooses SAOs to execute, as well as learning the APs for each SAO with convergence guarantees. We show the learned behaviours of the SAOs in a time dependent RoboCup soccer scenario and the Bottomless Pit of Death domain. \Alg\ has shown the potential to exit bad local optima due to sub-optimal reward shaping as well as mitigate feature-based model misspecification. In this work, we focus on the vigor awareness criterion. However, in principle, many other objectives can be incorporated into this framework that produce behaviours which include defence, aggression, risk and robustness  \citep{Avila1998,Tamar2015a,Tamar2015b,Leibo2017}. Extensions of this work include optimizing a PG-MDP performance threshold $\beta$ for each SAO as well as utilizing SA in lifelong learning problems \citep{Thrun1995,Brunskill2014}. It is also possible to implement the \Alg\ policy as a Deep Network \citep{Mnih2015}.\\
It has been argued that it is possible to model time-based SA using the regular expected utility by (1) performing reward shaping or (2) augmenting the state space with time. Automatically learning a SA strategy and determine when to use it, cannot be encoded into the reward function in a general, \textit{non-heuristic} manner and is subjective at most. In addition, the expected return has been shown to be inadequate for modelling SA behaviors such as risk (the most famous example being from \citep{kahneman1979prospect}). Augmenting the state space with time is both computationally inefficient and the Markovian assumption may not necessarily hold. 

\bibliography{tmann}
\bibliographystyle{icml2017}

\end{document}


\section{Full Derivation of Theorem 1}
We define the expected reward for following a policy $\mu_{\alpha,\Omega}$ as:

\begin{equation}
J(\mu_{\alpha, \Omega}) = \int_{\tau} P(\tau \vert \alpha, \Omega)R(\tau)d\tau \enspace .
\end{equation}

Let us group the inter-option policy parameters $\alpha \in \mathbb{R}^d$ and the continuous Awareness Distribution(AD) $\Omega \in \mathbb{R}^{|N||m|}$ parameters  into a single vector $\chi = [\alpha, \Omega] \in \mathbb{R}^{d+ |N||m|}$. 
Taking the derivative of this objective and using the well-known likelihood trick \cite{Peters2008} yields:

\begin{eqnarray}
\nabla_{\chi} J(\mu_{\chi}) &=& \nabla_{\chi} \int_{\tau} P(\tau \vert \chi)   R(\tau)d\tau\\
&=&  \int_{\tau}\nabla_{\chi} P(\tau \vert \chi)   R(\tau)d\tau\\
&=& \int_{\tau} P(\tau \vert \chi) \nabla_\chi \log P(\tau \vert \chi)  R(\tau) d\tau \enspace ,
\label{eqn:regPG}
\end{eqnarray}

where $P(\tau \vert \chi)=P(z_0)\prod_{k=0}^T P(z_{k+1} \vert z_k, o_k) P (o_k \vert z_k, \chi)$. Here, $z_k \in Z$ is the state at timestep $k$; $o_k$ is the SAS selected at timestep $k$ and $T$ is the length of the trajectory. Since only $P (o_k \vert z_k, \chi)$ is parameterized, the gradient $\nabla_\chi J(\mu_\chi)$ can be simplified as follows:

\begin{eqnarray}\nonumber
\nabla_{\chi} J(\mu_{\chi}) &=& \int_{\tau} P(\tau \vert \chi) \nabla_\chi \log P(\tau \vert \chi)  R(\tau)d\tau\\ \nonumber
&=& \int_{\tau} P(\tau \vert \chi) \nabla_\chi \log \biggl[P(z_0)\prod_{k=0}^T P(z_{k+1} \vert z_k, o_k) P (o_k \vert z_k, \chi) \biggr]  R(\tau)d\tau\\ \nonumber
&=& \int_{\tau} P(\tau \vert \chi) \nabla_\chi \biggl[ \log P(z_0)+\Sigma_{k=0}^T (\log P(z_{k+1} \vert z_k, o_k)+ \log P (o_k \vert z_k, \chi))  \biggr] R(\tau)d\tau\\ 
&=& \int_{\tau} P(\tau \vert \chi) \Sigma_{k=0}^T \nabla_\chi  \log P (o_k \vert z_k, \chi)  R(\tau)d\tau
\label{eqn:generalpg}
\end{eqnarray}

where $P (o_k \vert z_k, \chi) = \mu_{\alpha}(o_k|z_k)\mu_{\Omega}^{o_{k}}(c_k|z_k)$. Therefore, substituting the two-tiered policy into Equation \ref{eqn:generalpg} and deriving with respect to $\alpha$ leads to the gradient update rule:

\begin{eqnarray}\nonumber
\nabla_\alpha J(\mu_\chi) &=&  \int_{\tau} P(\tau \vert \chi)  \Sigma_{k=0}^T \nabla_\alpha \log \mu_{\alpha}(o_k|z_k) R(\tau)d\tau \enspace .
\label{eqn:generalpg1}
\end{eqnarray}

If we represent $\mu_{\alpha}(o_k|z_k)$ as a Gibb's distribution which is a common policy choice in many MDPs \cite{Sutton1998}, then we can easily estimate the gradient by sampling:

\begin{equation}
\nabla_\alpha J(\mu_\chi) = \left <  \sum_{k=0}^T  \nabla_\alpha \log  \mu_{\alpha}(o_k \vert z_k) \sum_{k=0}^T \gamma^{k} r_k     \right > \enspace ,
\end{equation}

where $H$ is the length of a trajectory; and $\biggl<\cdot \biggr>$ represents an average over trajectories. If we derive Equation \ref{eqn:generalpg} with respect to $\Omega$ for the AD parameters, then we get the following gradient update rule:

\begin{eqnarray}\nonumber
\nabla_\Omega J(\mu_\chi) &=&  \int_{\tau} P(\tau \vert \chi) \sum_{k=0}^T \nabla_\Omega \log \mu_{\Omega}^{o_{k}}(c_k|z_k) R(\tau)d\tau \enspace .
\label{eqn:generalpg2}
\end{eqnarray}

If we represent $\mu_{\Omega}^{o_{k}}(y_k|z_k)$ as any distribution from the natural exponential family, then we can easily estimate the gradient by samples using the following gradient update rule:

\begin{equation}
\nabla_\Omega J(\mu_\chi) = \left <  \sum_{k=0}^T  \nabla_\Omega \log  \mu_{\Omega}^{o_{k}}(c_k|z_k) \sum_{k=0}^T \gamma^k r_k     \right > \enspace .
\end{equation}

These derivations are summarized in Theorem $1$.

\begin{theorem}
Suppose that we are maximizing the Policy Gradient (PG) objective $J(\mu_{\alpha, \Omega}) = \int_{\tau} P(\tau\vert \alpha, \Omega)R(\tau)d\tau$ using awareness trajectories, generated by the two-tiered option selection policy $\mu_{\alpha}(\sigma_t|x_t)\mu_{\Omega}^{\sigma_{t}}(c_t|z_t)$, then the gradient update rules for the inter-option policy parameters $\alpha \in \mathbb{R}^d$ and the AD parameters $\Omega \in \mathbb{R}^{|N||m|}$ are defined in Equations \ref{eqn:grad1} and \ref{eqn:grad2} respectively.

\vspace{-0.6cm}
\begin{equation}
\nabla_\alpha J(\mu_{\alpha, \Omega}) = \left <  \sum_{h=0}^H  \nabla_\alpha \log  \mu_{\alpha}(\sigma_h|z_h) \sum_{j=0}^H \gamma^j r_j     \right >
\label{eqn:grad1}
\end{equation}
\begin{equation}
\nabla_\Omega J(\mu_{\alpha, \Omega}) = \left <  \sum_{h=0}^H  \nabla_\Omega \log  \mu_{\Omega}^{\sigma_{t}}(c_t|z_t) \sum_{j=0}^H \gamma^j r_j     \right > 
\label{eqn:grad2}
\end{equation}
$H$ is the trajectory length and $<\cdot >$ is an average over trajectories as in standard PG.
\label{thm:grad}
\end{theorem}

%
%

\section{Proof of Theorem 2: \Alg\ Convergence}

\begin{theorem}

Suppose we are optimizing the expected return $J(\mu_{\Omega,\alpha})=\intop P(\tau \vert \Omega,\alpha) R(\tau) d\tau$
for any arbitrary \Alg\ policy $\mu_{\Omega,\alpha}$ where $\alpha\in\mathbb{R}^{d}$ and $\Omega\in\mathbb{R}^{|N||m|}$ are the inter-option policy and AD parameters respectively. Then, for step sizes sequences
$\{a_{k}\}_{k=0}^{\infty},\{b_{k}\}_{k=0}^{\infty}$ that satisfy
$\sum_{k}a_{k}=\infty,\sum_{k}b_{k}=\infty,\sum_{k}a_{k}^{2}<\infty,\sum_{k}b_{k}^{2}<\infty$
and $b_{k}>a_{k}$, the \Alg\ iterates converge almost surely $\alpha_{k}\rightarrow\alpha^{*},\Omega_{k}\rightarrow\bar{\lambda}(\alpha^{*})$
to the countable set of locally optimal points of $J(\mu_{\Omega,\alpha})$
. 

\end{theorem}

The true gradient of the two-tiered policy $\mu_{\Omega,\alpha}$
is:

\[
\nabla_{\Omega,\alpha}J(\mu_{\Omega,\alpha})=\mathbb{E}[R_{\tau}\sum_{k=0}^T \nabla_{\Omega, \alpha} \log P(\tau \vert \Omega,\alpha)]
\]

where $R_{\tau}=\sum_{k=0}^{H}\gamma^{k}r_{k}$ is the discounted
cumulative reward for a trajectory $\tau$ of length $H$; the term
$P(\tau \vert \Omega,\alpha)=P(x_{0})\Pi_{k=0}^{H}P(x_{k+1}\vert x_{k},\sigma_{k}, c_k)\mu_{\Omega,\alpha}(\sigma_{k},c_{k}\vert x_{k})$
is the probability of a trajectory for a given policy $\mu_{\Omega,\alpha}(\sigma_{k},c_{k}\vert x_{k})$. 

The estimated gradient is:

\begin{eqnarray*}
\hat{\nabla}J(\mu_{\Omega, \alpha}) & = & R_{\tau} \sum_{k=0}^T   \nabla_{\Omega, \alpha} \log\mu_{\Omega,\alpha}(\sigma_{k},c_{k}\vert x_{k})\\
 & = & R_{\tau} \sum_{k=0}^T  \nabla_{\Omega, \alpha} \log(\mu_{\alpha}(\sigma_{k}\vert x_{k})\mu_{\Omega}^{\sigma_{k}}(c_{k}\vert x_{k}))
\end{eqnarray*}

We need to prove that the parameters $\alpha\in\mathbb{R}^{d}$ of
the inter-option policy and the AD parameters $\Omega\in\mathbb{R}^{|N||m|}$
converge to a locally optimal solution. Here, $N$ is the number of
options and $m$ is the number of AD parameters
for each option. In order to do so, we first derive the gradient with
respect to $\alpha$ to yield the following recursive update equations:

\begin{eqnarray}
\alpha_{k+1} & = & \Gamma_{\alpha}(\alpha_{k}+a_{k}\hat{\nabla}_{\alpha}J(x))\nonumber \\
 & = & \Gamma_{\alpha}(\alpha_{k}+a_{k}(R_{\tau} \sum_{k=0}^T  \nabla_{\alpha}\log\mu_{\alpha}(\sigma_{k}\vert x_{k})))\nonumber \\
 & \overset{(1)}{=} & \Gamma_{\alpha}(\alpha_{k}+a_{k}(R_{\tau}z_{k}^{\alpha}))\nonumber \\
 & = & \Gamma_{\alpha}(\alpha_{k}+a_{k}(R_{\tau}z_{k}^{\alpha}-\mathbb{E}[R_{\tau}z_{k}^{\alpha}]+\mathbb{E}[R_{\tau}z_{k}^{\alpha}]))\nonumber \\
 & = & \Gamma_{\alpha}(\alpha_{k}+a_{k}(f(\alpha(k),\Omega)+N_{k+1}))\label{eq:fast}
\end{eqnarray}
\\
where $(1)$ $z_{k}^{\alpha}=\sum_{k=0}^T \nabla_{\alpha}\log\mu_{\alpha}(\sigma_{k}\vert x_{k})$
and $N_{k+1}=R_{\tau}z_{k}^{\alpha}-\mathbb{E}[R_{\tau}z_{k}^{\alpha}]$
is a zero-mean martingale difference sequence; $f(\alpha(k),\Omega)=\mathbb{E}[R_{\tau}z_{k}^{\alpha}]$
and $\Gamma_{\alpha}:\mathbb{R}^{d}\rightarrow\mathbb{R}^{d}$ is
a projection operator that projects any $\alpha_{k}$ to a compact
region $C=\{\alpha\vert g_{i}(\alpha)\leq0,i=1,\cdots l\}\in\mathbb{R}^{n}$
where $g_{i}(\cdot),i=1,\cdots l$ represent the continuously differentiable
constraints that project the iterates to a compact region defined
by a ball with a smooth boundary. This operator ensures that the iterates
remain bounded. It can be seen by inspection that this recursion represents
a noisy discretization of the Ordinary Differential Equation (ODE)
\cite{Borkar1997}:

\[
\dot{\alpha}=\Gamma_{\alpha}(\mathbb{E}[R_{\tau}z_{k}^{\alpha}])
\]

We also derive the recursive update for the AD parameters
$\Omega$ as follows:

\begin{eqnarray}
\Omega_{k+1} & = & \Gamma_{\Omega}(\Omega_{k}+b_{k}\hat{\nabla}_{\Omega}J(x))\nonumber \\
 & = & \Gamma_{\Omega}(\Omega_{k}+b_{k}(R_{\tau} \sum_{k=0}^T  \nabla\log\mu_{\Omega}^{\sigma_{k}}(c_{k}\vert x_{k}))\nonumber \\
 & = & \Gamma_{\Omega}(\Omega_{k}+b_{k}(R_{\tau}z_{k}^{\Omega}))\nonumber \\
 & = & \Gamma_{\Omega}(\Omega_{k}+b_{k}(R_{\tau}z_{k}^{\Omega}-\mathbb{E}[R_{\tau}z_{k}^{\Omega}]+\mathbb{E}[R_{\tau}z_{k}^{\Omega}]))\nonumber \\
 & = & \Gamma_{\Omega}(\Omega_{k}+b_{k}(g(\Omega(k),\alpha)+M_{k+1})),\label{eq:slow}
\end{eqnarray}

where $M_{k+1}=R_{\tau}z_{k}^{\Omega}-\mathbb{E}[R_{\tau}z_{k}^{\Omega}]$
is a zero mean martingale difference sequence with respect to the
$\sigma-$fields $\mathcal{F}_{t}=\sigma(\Omega_{n},\alpha_{n},N_{n},M_{n},n\leq t;t\geq0)$;
$g(\Omega(k),\alpha)=\mathbb{E}[R_{\tau}z_{k}^{\Omega}]$ and $\Gamma_{\Omega}:\mathbb{R}^{|N||m|}\rightarrow\mathbb{R}^{|N||m|}$
is the corresponding projection operator for $\Omega$ which ensures
that these iterates are projected to a compact region $W$ as in the
previous iterate update equation. We can thus represent the $\Omega$
update with the following ODE:

\[
\dot{\Omega}=\Gamma_{\Omega}(\mathbb{E}[R_{\tau}z_{k}^{\Omega}])
\]
Define the continuous time projection operators $\hat{\Gamma}_{\Omega}(v)=\lim_{\delta\rightarrow\infty}\frac{\Gamma_{\Omega}(\Omega+\delta v)-\Omega}{\delta}$
and $\hat{\Gamma}_{\alpha}(p)=\lim_{\delta\rightarrow\infty}\frac{\Gamma_{\alpha}(\alpha+\delta v)-\alpha}{\delta}$
that, given directions $v$ and $p$ to modify the paramaters $\Omega$
and $\alpha$ respectively ensures that the iterates are projected
into their compact sets $C$ and $W$ respectively. We can thus define
the ODEs using this continuous operator as:

\begin{align*}
\dot{\Omega} & =\hat{\Gamma}_{\Omega}(\mathbb{E}[R_{\tau}z_{k}^{\Omega}])\doteq\bar{g}(\Omega(k),\alpha)\\
\dot{\alpha} & =\hat{\Gamma}_{\alpha}(\mathbb{E}[R_{\tau}z_{k}^{\alpha}])\doteq\bar{f}(\alpha(k),\Omega)
\end{align*}
\textbf{Assumption (A1): }For each $\alpha\in\mathbb{R}^{d}$, the
ODE:

\[
\dot{\Omega}(k)=\bar{g}(\Omega(k),\alpha)
\]

has a globally asymptotically stable equilibrium $\bar{\lambda}(\alpha)$
such that $\bar{\lambda}:\mathbb{R}^{d}\rightarrow\mathbb{R}^{|N||m|}$
is Lipschitz.

\textbf{Assumption (A2): }The ODE:

\[
\dot{\alpha}=\bar{f}(\alpha(k),\bar{\lambda}(\alpha(k)))
\]

has a unique global asymptotically stable equilibrium $\alpha^{*}$.

In order to prove that these ODEs collectively converge, we need to
make the following assumptions.

\textbf{Assumption (A3): }The functions $f,g$ are Lipschitz continuous
functions

\textbf{Assumption (A4): }$\sup_{k}\Vert\Omega_{k}\Vert,\sup_{k}\Vert\alpha_{k}\Vert<\infty$

\textbf{Assumption (A5): }$\sum_{n}a(n)=\sum_{n}b(n)=\infty,$\textbf{
$\sum_{n}a(n)^{2}=\sum_{n}b(n)^{2}<\infty$}

\textbf{Assumption (A6): }For increasing $\sigma$-algebras, the martingale
sequences $\sum a_{k}N_{k},\sum b_{k}M_{k}<\infty$ a.s 

\textbf{Assumption (A7) }For all $\alpha,\Omega$, the objective function
$J(\mu_{\alpha,\Omega})$ has bounded second derivatives and the set
$Z$ of local optima of $J(\mu_{\alpha,\Omega})$ are countable.

Given the above assumptions, the parameter $\Omega_{k}\rightarrow\bar{\lambda}(\alpha^{*})$
and $\alpha_{k}\rightarrow\alpha^{*}$ as $k\rightarrow\infty$ a.s
by standard two-timescale stochastic approximation arguments \cite{Borkar1997}.
That is, the iterates converge to \{$\bar{\lambda}(\alpha^{*}),\alpha^{*}\vert\alpha^{*}\in Z\}$
.

\bibliography{tmann}
\bibliographystyle{icml2017}